\documentclass[conference]{IEEEtran}
\IEEEoverridecommandlockouts
\usepackage{caption} % 添加标题
\usepackage{cite}
\usepackage{hyperref} % 用于跳转到引用

\usepackage{booktabs} % 表格样式
\usepackage{multirow} % 表格样式，引入multirow宏包
\usepackage{diagbox} % 表格样式，表格斜线

\usepackage{amsmath,amssymb,amsfonts}
\usepackage{graphicx}
\usepackage{textcomp}
\usepackage{color}
\usepackage[dvipsnames]{xcolor}

\def\BibTeX{{\rm B\kern-.05em{\sc i\kern-.025em b}\kern-.08em
    T\kern-.1667em\lower.7ex\hbox{E}\kern-.125emX}}
    
\usepackage{algpseudocode} % 伪代码样式 !导致产生1.5em 0pt
\usepackage{algorithm} % 伪代码样式

\usepackage{flushend} % 参考文献双栏平均
\usepackage[numbers, sort]{natbib} % 保证参考文献不会出现乱序
\begin{document}

\title{\textcolor{cyan}{I}\textcolor{magenta}{O}\textcolor{PineGreen}{V}\textcolor{orange}{S}4NeRF: \textcolor{cyan}{I}ncremental \textcolor{magenta}{O}ptimal \textcolor{PineGreen}{V}iew \textcolor{orange}{S}election\\ for Large-Scale NeRFs\\
\thanks{This work is supported by the National Key R\&D Program of China (Grant NO. 2022ZD0119000), Hunan Provincial Key R\&D Program of China (Grant NO. 2024JK2020 and 024JK2021), Hunan Provincial Natural Science Foundation of China (Grant NO. 2024JJ10027), Young Talents of Huxiang (Grant NO. Z202433000575), Lushan Lab Research Funding, and Changsha Science Fund for Distinguished Young Scholars (Grant NO. kq2306002).}
}

\author{\IEEEauthorblockN{
\textit{Jingpeng Xie}\textsuperscript{1,$\dagger$},
\textit{Shiyu Tan}\textsuperscript{1,$\dagger$},
\textit{Yuanlei Wang}\textsuperscript{1},
\textit{Tianle Du}\textsuperscript{2},
\textit{Yifei Xue}\textsuperscript{1,$\ast$},
\textit{Yizhen Lao}\textsuperscript{1,$\ast$}
}\thanks{$\dagger$ Equal contribution. $\ast$ Corresponding authors.}
\\
\IEEEauthorblockA{
\textsuperscript{1}Hunan University, Changsha, China \\
\textsuperscript{2}Nanchang University, Nanchang, China}
}
\maketitle

\begin{abstract}
Large-scale Neural Radiance Fields (NeRF) reconstructions are typically hindered by the requirement for extensive image datasets and substantial computational resources. This paper introduces IOVS4NeRF, a framework that employs an uncertainty-guided incremental optimal view selection strategy adaptable to various NeRF implementations. Specifically, by leveraging a hybrid uncertainty model that combines rendering and positional uncertainties, the proposed method calculates the most informative view from among the candidates, thereby enabling incremental optimization of scene reconstruction. Our detailed experiments demonstrate that IOVS4NeRF achieves high-fidelity NeRF reconstruction with minimal computational resources, making it suitable for large-scale scene applications.
\end{abstract}

\begin{IEEEkeywords}
NeRF, Uncertainty Estimation, View Selection.
\end{IEEEkeywords}

\section{Introduction}
%% Introduction 第一段(√)
%Neural Radiation Fields (NeRF)\cite{mildenhall2021nerf} is a novel view synthesis method that combines 3D reconstruction and neural rendering. Due to implicit scene representation and differentiable volume rendering\cite{yu2021pixelnerf}, its reconstruction is low-efficiency. Previous works\cite{muller2022instant, chen2022tensorf, fridovich2022plenoxels, sun2022direct, xu2022point} utilize combinations of hash functions, spherical harmonics, and both explicit and implicit scene representations to enable rapid training of NeRF. However, it still cannot cope with the challenge of large data volumes in large-scale scenes. Current NeRF-based methods\cite{turki2022mega, tancik2022block, zhenxing2022switch} for large-scale scenes mainly utilize the idea of decomposing the large scene into several small scenes for training, and are built on the premise of consuming a large amount of resources. Only meeting limited image inputs will result in severe artifacts and lower visual fidelity in the reconstructed results.

Neural Radiance Fields (NeRF)~\cite{mildenhall2021nerf} is an innovative method for synthesizing views by integrating 3D reconstruction with neural rendering. It uses implicit scene representation and differentiable volume rendering~\cite{yu2021pixelnerf}, requiring significant computational resources for large-scale scene reconstruction, often leading to inefficient processes~\cite{mildenhall2021nerf}.
Previous works~\cite{muller2022instant, chen2022tensorf, fridovich2022plenoxels, sun2022direct, xu2022point} utilize combinations of hash functions, spherical harmonics, and both explicit and implicit scene representations to enable rapid training of NeRF. 
However, it still cannot cope with the challenge of large-scale scenes.
Current efforts~\cite{turki2022mega, tancik2022block, zhenxing2022switch} have been made by decomposing large scenes into smaller segments, each trained by an isolated NeRF submodule, which necessitating substantial computational resources.
In resource-limited settings, reconstructions exhibit severe artifacts and diminished visual fidelity in large scenes.

%% Introduction 第二段（√）
%Although NeRF is effective, its performance is greatly affected by the quality of the training samples.
%Inspired by the Next-Best View (NBV) problem~\cite{zeng2020view}, we expect each image in the limited input to contain as much information as possible to maximize the understanding of the scene and solve the problem of training large-scale NeRF under resource constraints. Emerging works~\cite{shen2021stochastic, shen2022conditional, pan2022activenerf} use uncertainty-guided NBV selection but often alter the internal network, impacting reusability. Other works~\cite{sunderhauf2023density, lee2022uncertainty} don't make any changes to NeRF itself, but mostly only consider pixel-level uncertainty during the rendering process. These methods have achieved good results, but still face challenges with artifacts and efficiency when applied to large-scale scenes.
Drawing inspiration from the Next-Best View (NBV) problem~\cite{zeng2020view}, we expect each image in the limited input set to contain as much information as possible, thereby maximizing scene understanding and addressing the challenge of training large-scale NeRF under resource constraints. Emerging studies~\cite{shen2021stochastic, shen2022conditional, pan2022activenerf} use uncertainty-guided NBV selection, but they often modify the internal network, which can affect reusability. Other approaches~\cite{sunderhauf2023density, lee2022uncertainty} make no changes to NeRF architecture but focus primarily on pixel-level uncertainty during rendering. While these methods have shown promise, they still face issues with artifacts and efficiency when applied to large-scale scenes.

%% 图一（√）
\begin{figure}[!htbp]
    %% \setlength{\abovecaptionskip}{-0.2cm} %% 调整标题与图之间的间距
    %% \captionsetup{font=footnotesize}
	\centering
	\includegraphics[width=1.0\linewidth]{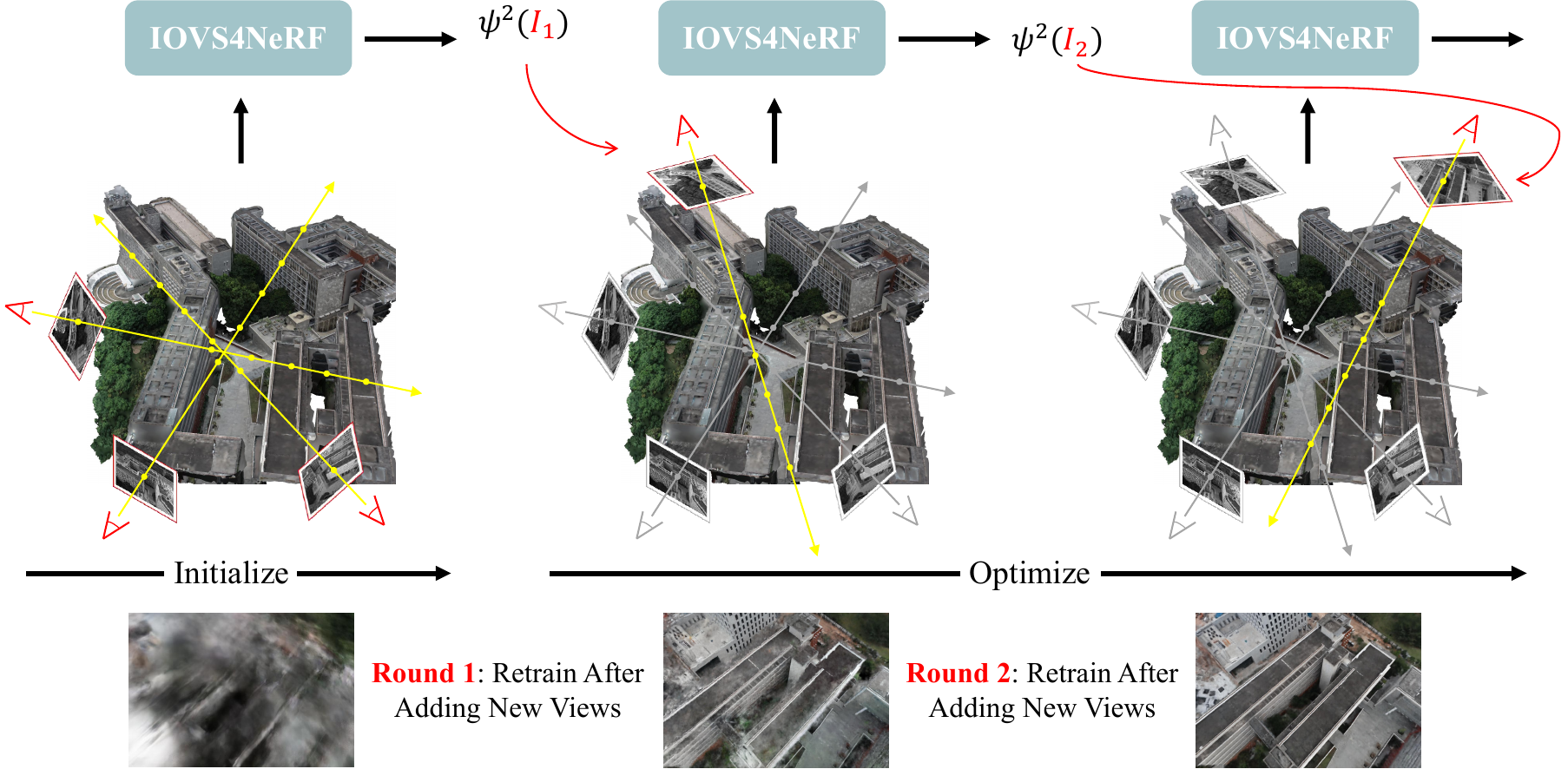}
	\caption{\textbf{IOVS4NeRF} is a flexible framework that actively expands the existing training set with newly captured samples based on hybrid uncertainties of candidate views.}
    \label{iovs4}
    \vspace{-4mm}
\end{figure}

%% 图二（√）
\begin{figure*}
	\centering
	\textbf{}
    \includegraphics[width=0.8\linewidth]{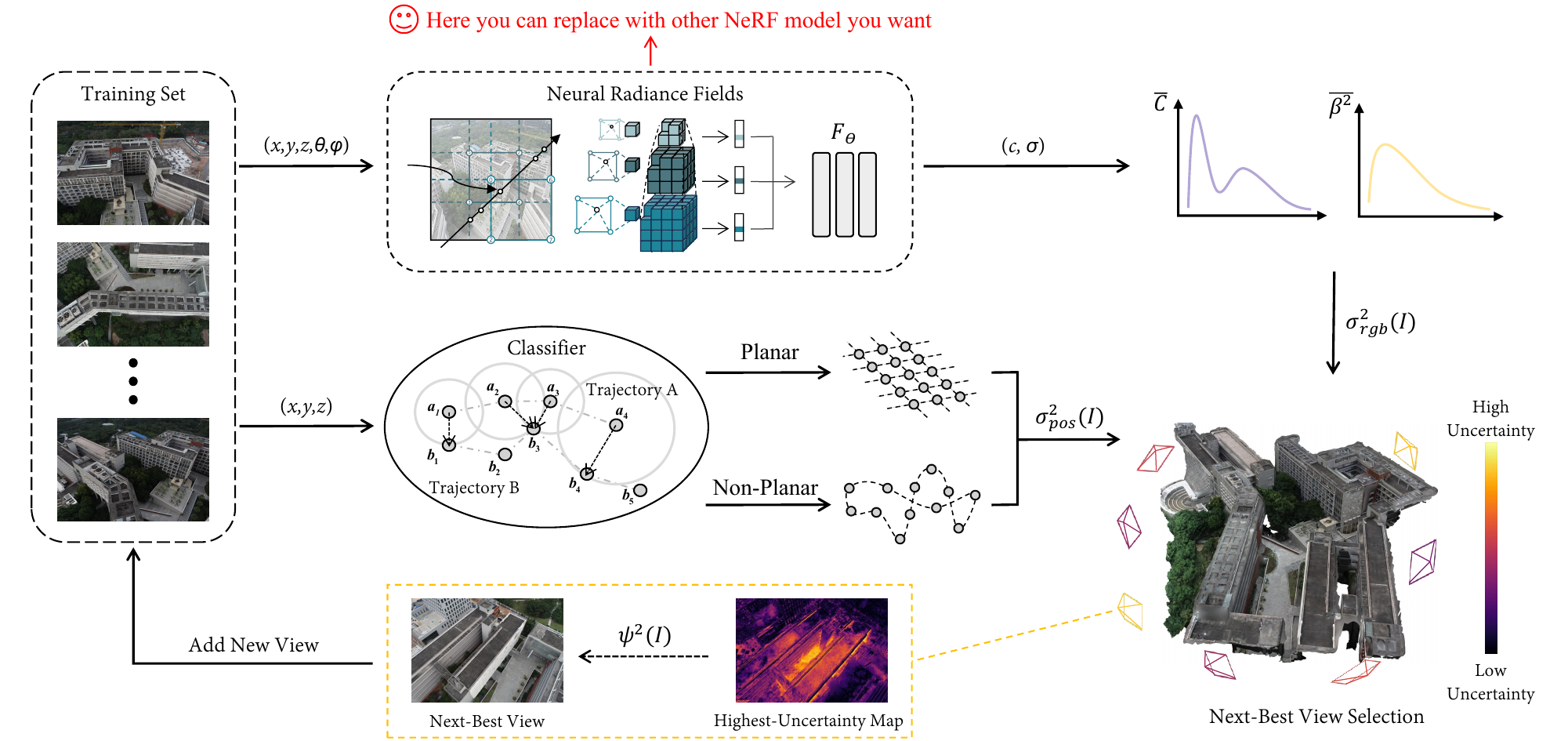}
	\caption{\textbf{Overview of Our Framework.} It consists of three parts: I. We train NeRF using 5D image information, modeling color as a Gaussian distribution to compute rendering uncertainty. II. We calculate positional uncertainty from training images using a classifier, combining it with rendering uncertainty to form hybrid uncertainty. III. The image with the highest hybrid uncertainty is iteratively added to the training set until the desired reconstruction quality or image limit is reached.}
    \label{fig:framework}
    \vspace{-2mm}
\end{figure*}

%% Introduction 第三段（√）
Building on the above background, we model the emission radiation values of each position in the scene as a Gaussian distribution rather than a single value. This approach prevents NeRF from collapsing into trivial solutions with incomplete scene observations. Inspired by oblique photography~\cite{zhou2021application}, we incorporate both planar and non-planar UAV flight paths~\cite{roberts2017submodular, petrie2009systematic} to assess positional uncertainty. We combine these uncertainties into a hybrid metric to optimize view selection, as illustrated in Fig.~\ref{iovs4}.

%% Introduction 第四段（√）
Furthermore, we accelerate the reconstruction speed by employing a lightweight NeRF from the perspective of the model structure. In this paper, we use Instant-NGP~\cite{muller2022instant} because its multi-resolution hash encoding has adaptive and efficient characteristics, making it well-adaptable to scenes of different scales. In summary, our contributions are as follows:

%% Introduction 第五段（√）
\begin{itemize}
\item We propose a NeRF-based framework for incremental optimal view selection to address the limitations of NeRFs in large-scale scenes with restrained computational cost.
\item We define a novel estimation of uncertainty, consisting of rendering and positional uncertainty, to compute the information gain of each candidate view.
\item We evaluate our method on real-world datasets, showing it significantly reduces time while maintaining reconstruction effectiveness compared to previous baselines.
%\item The code will be \textit{publicly available} upon acceptance of the paper
\end{itemize}

\section{Methodology}
%% Methodology 第一段（√）
As illustrated in Fig.~\ref{fig:framework}, our proposed IOVS4NeRF introduces, in epistemic uncertainty mind~\cite{kendall2017uncertainties}, a hybrid uncertainty, which can be reduced by collecting more data.

%\subsection{Hybrid Uncertainty}\label{AA}
%% Methodology 第二段（√）
IOVS4NeRF selects the best view using a \textit{hybrid uncertainty} that combines rendering and positional uncertainty. For rendering uncertainty, 5D coordinates $(\textbf{x},\textbf{d})$ are input into a NeRF network, which outputs color $c$ and volume density $\sigma$. These are used to calculate image uncertainty as a Gaussian distribution. Positional uncertainty is derived using 3D coordinates \textbf{$x$}, with a classifier determining trajectory type and Voronoi diagrams estimating information. These two uncertainties are normalized and summed to form the hybrid uncertainty:
\vspace{-1mm}

%% 公式一（√）
\begin{equation} 
    \label{eq:hybrid-uncertainty-definition}
	{\psi ^2}(I) = Norm(\sigma _{pos}^2(I)) + Norm(\sigma _{rgb}^2(I)),
\end{equation} where \textit{I} is the ground truth image, and $\sigma _{rgb}^2(I)$ and $\sigma _{pos}^2(I)$ are rendering and positional uncertainty respectively. 

%% Methodology 第三段（√）
\subsection{Rendering Uncertainty}
NeRF~\cite{mildenhall2021nerf} represents a scene as a continuous function \( F_\theta \) that outputs both the emitted radiance value and volume density. Specifically, given a 3D  \(\mathbf{x} = (x, y, z)\) and a viewing direction vector \(\mathbf{d} = (\theta,\phi)\), a multi-layer perceptron model is used to generate the corresponding color $c$ and volume density $\sigma$.

%% Methodology 第四段（√）
NeRF utilizes volume rendering to determine the color of rays passing through the scene, enabling novel view synthesis. Given a camera ray \(\mathbf{r}(t) = \mathbf{o} + t\mathbf{d}\) with the camera center \(\mathbf{o} \in \mathbb{R}^3\) passing through a specific pixel on the image plane, the color of the pixel is calculated by volume rendering integration, approximating integrals via stratified sampling discrete points can expressed as:
\vspace{-1mm}

%% 公式二（√）	
\begin{equation}
    \label{eq:2}
	\begin{aligned}
	\hat{C}(\mathbf{r}) &= \sum_{i=1}^{N_s} \alpha_i c(\mathbf{r}(t_i)),
	\end{aligned}
\end{equation}

%% 公式三（√）
\begin{equation}
    \label{eq:3}
	\begin{aligned}
	\alpha_i &= \exp\left(-\sum_{j=1}^{i-1} \sigma_j \delta_j\right) \left(1 - \exp(-\sigma_i \delta_i)\right) ,
	\end{aligned}
\end{equation} where \(\delta_i = t_{i+1} - t_i\) is the distance between samples, and \(N_s\) denotes the number of samples.

%% Methodology 第五段（√）
To improve results with limited training data, we model RGB values as Gaussian distributions\cite{pan2022activenerf}, using variance to represent uncertainty. We add a variance branch to NeRF's MLP, defining variance via information entropy:
\vspace{-1mm}

%% 公式四（√）
\begin{equation} 
    \label{eq:variance-def}
	{\beta ^2}({\bf{r}}({t_i})) =  - P({\alpha _i})\log P({\alpha _i}),
\end{equation} where ${\alpha _i}$ is the same as Eq.~(\ref{eq:2}), and $P( \cdot )$ represents proportion of ${\alpha _i} $in the $\sum {{\alpha _i}} $ along the ray.

%% Methodology 第六段（√）
Given the conjugacy of Gaussian distributions, the rendered values along a ray also follow a Gaussian distribution:
\vspace{-1mm}

%% 公式五（√）
\begin{equation} 
\label{eq:c-hat-def}
	\hat C({\textbf{r}}) \sim Beta(\bar C(\textbf{r}),{{\bar \beta }^2}(\textbf{r})),
    \vspace{-1mm}
\end{equation} where 

%% 公式六（√）
\begin{equation} 
\label{eq:c-bar-def1}
	\bar C({\textbf{r}}) = \sum\limits_{i = 1}^{{N_s}} {{\alpha _i}c({\textbf{r}}({t_i}))},   
     {\bar \beta ^2}({\textbf{r}}) = \sum\limits_{i = 1}^{{N_s}} {\alpha _i^2{\beta ^2}({\textbf{r}}({t_i}))} )
\end{equation}

%% 公式七（√）
Assuming ray independence, rendering uncertainty is maximized by the log-likelihood of rays from the same image:
\begin{equation} 
\label{eq:rgb-def}
	\sigma _{rgb}^2(I) = \sum\limits_{i = 1}^{{N_r}} {\frac{{||C({{\bf{r}}_i}) - \bar C({{\bf{r}}_i})||_2^2}}{{2{{\bar \beta  }^2}({{\bf{r}}_i})}} + \frac{{\log {{\bar \beta  }^2}({{\bf{r}}_i})}}{2}}, 
\end{equation}	where $N_{r}$ means the total number of the rays in photo $\textsl{I}$. 

%% 图三（√）
\begin{figure}
\label{fig:f3}
	\centering
	\includegraphics[width=1.0\linewidth]{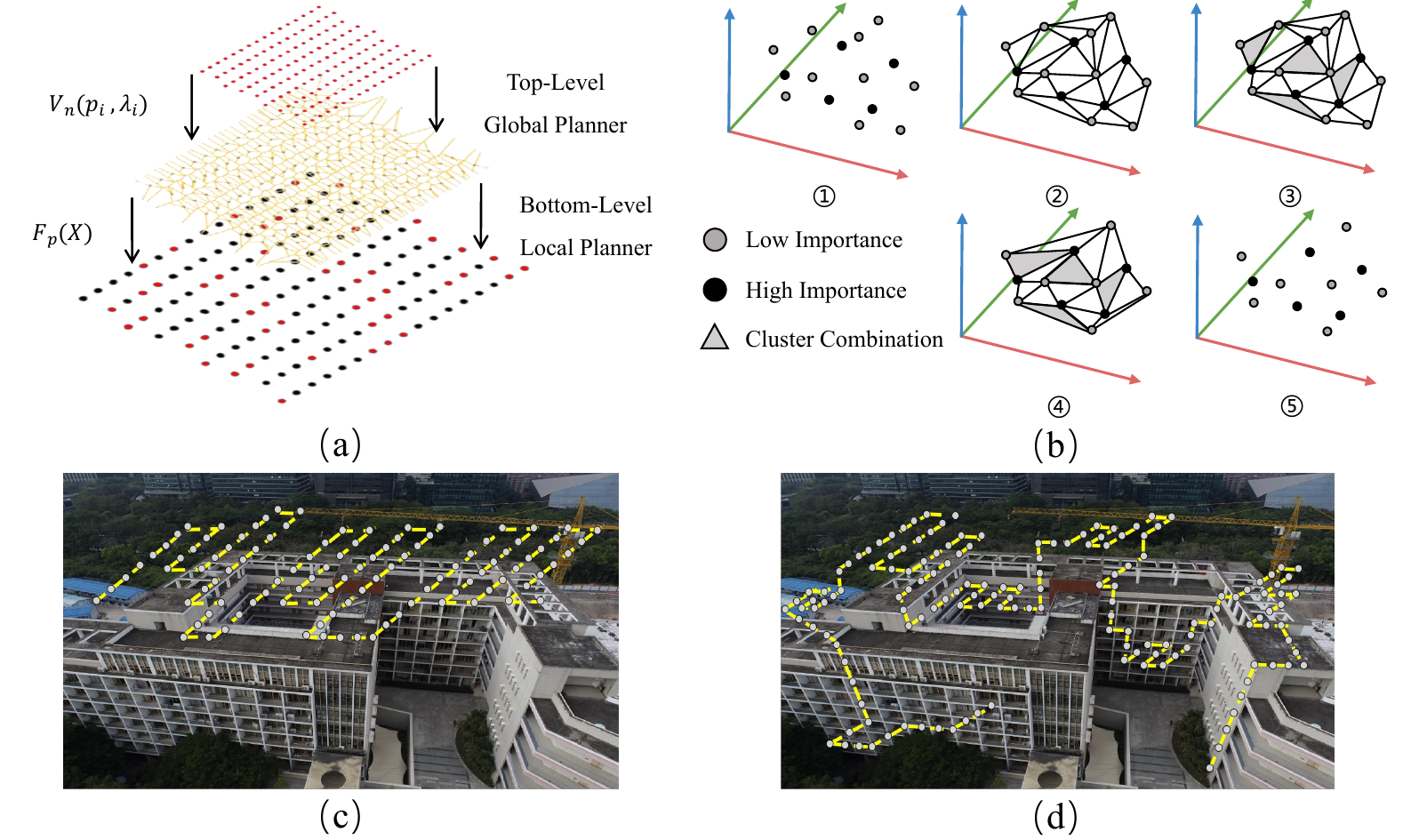}
	\caption{(a) We use a Voronoi-based method, where a top-level planner updates nodes and a bottom-level planner refines them based on potential fields. (b) We cluster points based on maximum volume and local density, quantifying uncertainty through importance values and density measures. (c) Planar flight trajectory. (d) Non-planar flight trajectory. }
    \vspace{-4mm}
\end{figure}

\subsection{Positional Uncertainty}
%% Methodology 第七段（√）
It has been determined that most UAV flight paths that can be programmed follow an approximate planar trajectory\cite{zhou2021application}, but users can freely manipulate the UAV to fly along non-planar trajectories. Therefore, we design a classifier to determine whether a UAV flight trajectory is planar or non-planar. To implement this adaptive classifier, we divide the data points into two groups, \(A = \{a_1, \ldots, a_p\}\) and \(B = \{b_1, \ldots, b_q\}\), and measure their similarity using the Hausdorff distance\cite{dubuisson1994modified}.
The Hausdorff distance offers a direct measure of trajectory similarity, bypassing interpolation and reducing noise. We set a threshold and classify trajectories as non-planar if their distance exceeds this threshold.

%% Methodology 第八段（√）
For planar trajectories, we use a Voronoi-based information gain radiance field\cite{ok2013path} within a hierarchical planning framework. The top-level global planner creates a collision-free graph using an improved Voronoi diagram. It traverses nodes to cover maximum uncertainty, while the bottom-level local planner refines this by selecting the most significant node based on weighted centroids. Specifically, we compress 3D pose information points to two-dimensional Euclidean space and define ${p_i} $ as n distinct points on this space, and $\lambda_i$ as the weighted value of a given point. Then, $% 
{{\mathop{\rm V}\nolimits} _n}({p_i},{\lambda _i})$ is the V-region of point ${p_i} $ with weight $\lambda_{i}$ , where ${d(p,{p_i})}$ is the Euclidean distance between ${p} $ and ${p_i} $:

%% 公式八（√）
\begin{equation} \label{eq:v-region-def}
	{{\mathop{\rm V}\nolimits} _n}({p_i},{\lambda _i}) = \bigcap\limits_{j \ne i} {\left\{ {p|\frac{{d(p,{p_i})}}{{{\lambda _i}}} < \frac{{d(p,{p_j})}}{{{\lambda _j}}}} \right\}}
\end{equation} 

In a non-confusing context, we abbreviate $% 
{{\mathop{\rm V}\nolimits} _n}({p_i},{\lambda _i})$ as 
${{{\mathop{\rm V}\nolimits} _n}({p_i})}$. Based on this, the uncertainty of our plane's Voronoi regions is defined as:
\vspace{-2mm}

%% 公式九（√）
\begin{equation} 
\label{eq:fp-def}
	{F_p}(I) = \sum\limits_{i = 1}^{{N_v}} {\frac{{\sum\limits_{j = 1}^{{N_v}} {||{p_i} - {p_j}|{|^{{\lambda _i}}}} }}{{{A_i}}}}, 
\end{equation} where ${{A_i}}$ represents area of the Voronoi polygon ${{V_n}({p_i})}$, and ${{N_v}}$ means the total number of 3D pose points.

%% Methodology 第九段（√）
We use the Voronoi clustering algorithm\cite{yan2008algorithm} to cluster points based on a volume threshold for non-planar trajectories. Point cluster generalization translates large-scale information into smaller diagrams, considering topological and metric factors. We combine importance values and local density to evaluate changes in importance across areas. The larger the area of the Voronoi polygon, the greater the weight. Thus, points are more likely to be retained in generalized mapping. The importance value equation is as follows:
\vspace{-1mm}

%% 公式十（√）
\begin{equation}
\label{eq:gi-def}
	{G_i} = \frac{{{\lambda _i}{A_i}}}{{\sum\limits_{k = 1}^{{N_v}} {({\lambda _k}{A_k})} }},
\end{equation}
\noindent where $G_{i}$ is the probability of selecting the specific point and given by the product of the area of the Voronoi polygon, denoted as $A_{i}$, and the weight value, denoted as $\lambda_{i}$.

%% Methodology 第十段（√）
Relative local density enables the comparison of density changes point by point before and after generalization, thus better assessing the density changes between points before and after generalization. As follows:
\vspace{-1mm}

%% 公式十一（√）
\begin{equation}
\label{eq:ri-def}
	{r_i} = \frac{{\frac{1}{{{A_i}}}}}{{\sum\limits_{k = 1}^{{N_v}} {\frac{1}{{{A_k}}}} }}
\end{equation}

The positional uncertainty of non-planer trajectory is:
\vspace{-2mm}

%% 公式十二（√）
\begin{equation} 
\label{eq:fnp-def}
	{F_{np}}(I) = \sum\limits_{i = 1}^{{N_v}} { - \log ({G_i}){r_i} + {\lambda _i}{{\left\| {{G_i} - {r_i}} \right\|}^2}}
\end{equation}

%% 公式十三（√）
Therefore, the total positional uncertainty is:
\begin{equation} 
\label{eq:pos-def}
    \sigma_{pos}^2(I) = \mathbb{I}(p) F_p(I) + \mathbb{I}(np) F_{np}(I),
\end{equation} where \( \mathbb{I}( \cdot ) \) is the indicator function.

\section{Experiments}
\subsection{Experimental Setup}\label{SCM}
%% Experiments第一段（√）
The six sets of UAV data we use come from four large-scale datasets: two benchmarks from Mill-19, two from Urban-3D, one from Pix-4D, and one self-captured footage dataset from Changsha, Hunan. Due to GPU limitations, we randomly select 500 images for each dataset as the whole image set (if the total number of images in a dataset is less than 500, we use all the images instead). We randomly choose 15\% of the images in each dataset for initialization and randomly select 10\% of the images as the test set. Then, we incrementally select 15\% of the images using various view selection methods as the optimal views for incremental training. Thus, only 30\% of the full image set is used to train.

%% Experiments第二段（√）
We input the ground truth RGB images into COLMAP, which is a software to recover the structures from images to get the pose of each image and utilize the Instant-NGP\cite{muller2022instant} to demonstrate the effectiveness of the proposed IOVS4NeRF. All the experiments are conducted with an Intel core I9 CPU and an NVIDIA GeForce 3090 GPU (24GB memory). 

%% 图四（√）
\begin{figure}
	\centering
	\includegraphics[width=1.0\linewidth]{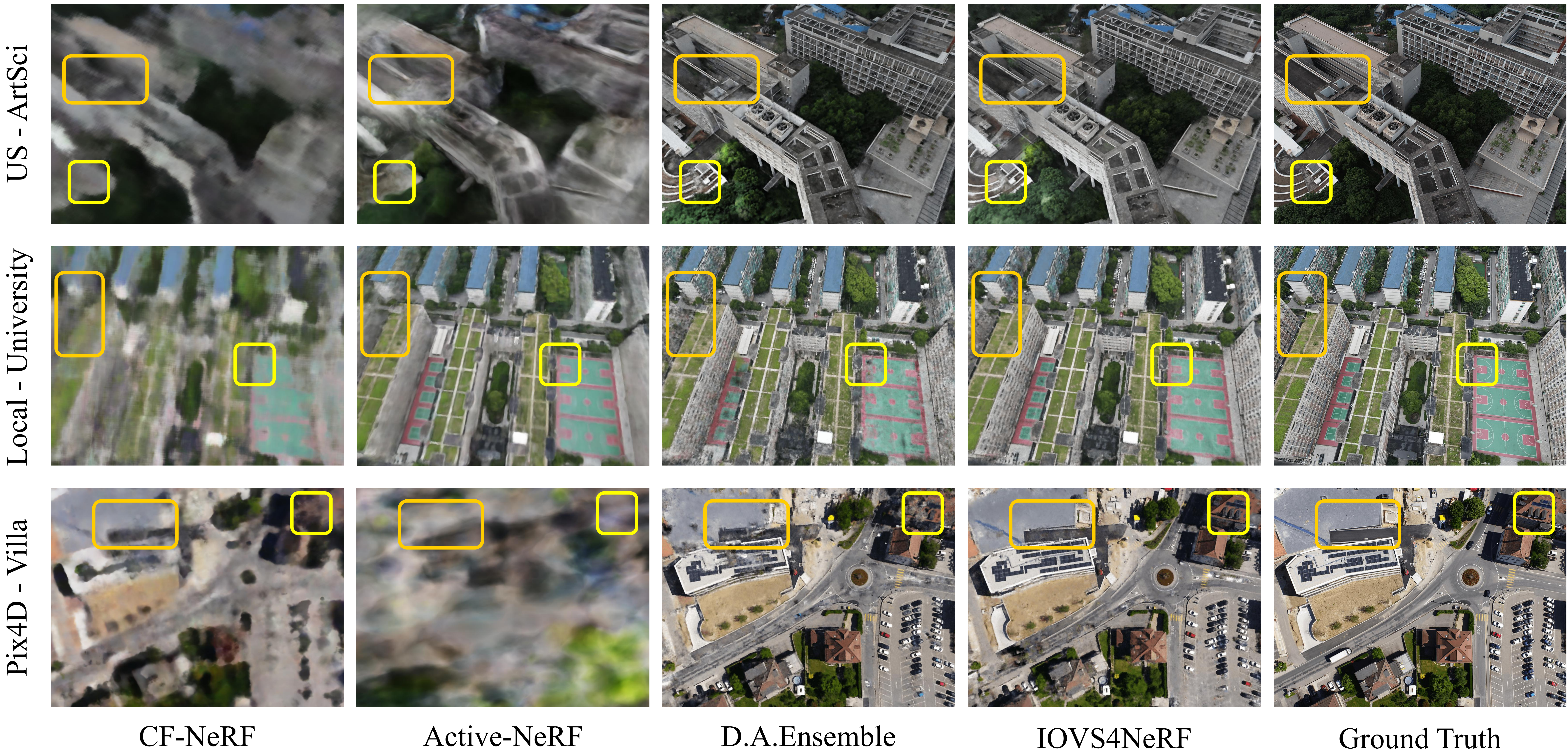}
	\caption{Rendering effect visualization of different methods (partial experimental results).}
	\label{fig:real-experiment}
    \vspace{-2mm}
\end{figure}

\subsection{Comparison Experiment}\label{AAA}
%% Experiments第三段（√）
To evaluate the results and computational costs of methods applied to large-scale scene reconstruction of UAV, we will compare IOVS4NeRF with three current popular methods\cite{shen2022conditional, pan2022activenerf, sunderhauf2023density} designed to exclusively estimate uncertainty in NeRF-based models. Additionally, evaluation metrics are considered from both image quality and uncertainty.

%% 表格一（√）
\begin{table}[ht]
	\setlength{\tabcolsep}{1pt} % 设置表格列之间的间距为 1pt
	\caption{Quantitative results of different methods in comparison experiment (best shown in \textcolor{green}{green} and second in \textcolor{blue}{blue}).}
	\begin{tabular}{cccccccc}
		\toprule
		\multirow{2}{*}{Scenes} & \multirow{2}{*}{Methods}  & \multicolumn{3}{c}{Quality} & \multicolumn{2}{c}{Uncertainty} & \multirow{2}{*}{Time↓}\\
		\cmidrule{3-5} \cmidrule(l){6-7}
		& & PSNR↑ & SSIM↑ & LPIPS↓  & AUSE↓ & SRCC↑ \\
		\midrule
		\multirow{4}{*}{Building} & CF-NeRF & 14.926 & 0.260 & 0.738 & \textcolor{blue}{0.100} & \textcolor{blue}{0.706} & 12:57 \\
		& Active-NeRF & 12.430 & 0.231 & 0.777 & 0.141 & 0.464 & 17:45 \\
		& D.A.Ensemble & \textcolor{blue}{16.857} & \textcolor{blue}{0.383} & \textcolor{blue}{0.422} & 0.303 & 0.668 & \textcolor{blue}{12:18} \\
		& IOVS4NeRF & \textcolor{green}{19.841} & \textcolor{green}{0.469} & \textcolor{green}{0.351} & \textcolor{green}{0.055} & \textcolor{green}{0.878} & \textcolor{green}{7:28} \\
		\midrule
		\multirow{4}{*}{Rubble} & CF-NeRF & 17.758 & 0.312 & 0.717 & \textcolor{blue}{0.075} & \textcolor{blue}{0.815} & 12:58 \\
		& Active-NeRF & 16.531 & 0.275 & 0.622 & 0.097 & 0.683 & 17:47 \\
		& D.A.Ensemble & \textcolor{blue}{19.629} & \textcolor{blue}{0.507} & \textcolor{blue}{0.335} & 0.257 & \textcolor{blue}{0.818} & \textcolor{blue}{12:20} \\
		& IOVS4NeRF & \textcolor{green}{21.108} & \textcolor{green}{0.573} & \textcolor{green}{0.303} & \textcolor{green}{0.052} & \textcolor{green}{0.884} & \textcolor{green}{7:29} \\
		\midrule
		\multirow{4}{*}{Polytech} & CF-NeRF & 14.259 & 0.251 & 0.731 & 0.118 & \textcolor{blue}{0.751} & 12:57 \\
		& Active-NeRF & 8.888 & 0.153 & 0.747 & 0.264 & -0.035 & 17:46 \\
		& D.A.Ensemble & \textcolor{blue}{20.673} & \textcolor{blue}{0.564} & \textcolor{blue}{0.253} & \textcolor{blue}{0.058} & \textcolor{blue}{0.915} & \textcolor{blue}{12:46} \\
		& IOVS4NeRF & \textcolor{green}{21.810} & \textcolor{green}{0.592} & \textcolor{green}{0.235} & \textcolor{green}{0.053} & \textcolor{green}{0.925} & \textcolor{green}{7:48} \\
		\midrule
		\multirow{4}{*}{ArtSci} & CF-NeRF & 15.578 & 0.227 & 0.693 & \textcolor{blue}{0.106} & 0.729 & 12:57 \\
		& Active-NeRF & 13.531 & 0.193 & 0.610 & 0.152 & 0.525 & 17:48 \\
		& D.A.Ensemble & \textcolor{blue}{17.769} & \textcolor{blue}{0.463} & \textcolor{blue}{0.338} & 0.101 & \textcolor{blue}{0.793} & 13:28 \\
		& IOVS4NeRF & \textcolor{green}{18.836} & \textcolor{green}{0.500} & \textcolor{green}{0.319} & \textcolor{green}{0.083} & \textcolor{green}{0.834} & \textcolor{green}{8:21} \\
		\midrule
		\multirow{4}{*}{University} & CF-NeRF & 15.359 & 0.189 & 0.669 & \textcolor{blue}{0.112} & 0.645 & 12:58 \\
		& Active-NeRF & \textcolor{blue}{17.814} & \textcolor{blue}{0.359} & \textcolor{blue}{0.359} & 0.087 & \textcolor{blue}{0.780} & 17:46 \\
		& D.A.Ensemble & 16.610 & 0.331 & 0.416 & 0.101 & 0.682 & \textcolor{blue}{12:08} \\
		& IOVS4NeRF & \textcolor{green}{18.722} & \textcolor{green}{0.430} & \textcolor{green}{0.327} & \textcolor{green}{0.077} & \textcolor{green}{0.814} & \textcolor{green}{6:47} \\
		\midrule
		\multirow{4}{*}{Villa} & CF-NeRF & 15.141 & 0.259 & 0.645 & \textcolor{blue}{0.111} & \textcolor{blue}{0.777} & \textcolor{blue}{12:58} \\
		& Active-NeRF & 9.236 & 0.121 & 0.828 & 0.242 & 0.055 & 20:32 \\
		& D.A.Ensemble & \textcolor{blue}{15.415} & \textcolor{blue}{0.337} & \textcolor{blue}{0.494} & 0.123 & 0.689 & 13:58 \\
		& IOVS4NeRF & \textcolor{green}{18.250} & \textcolor{green}{0.457} & \textcolor{green}{0.386} & \textcolor{green}{0.079} & \textcolor{green}{0.868} & \textcolor{green}{8:46} \\
		\bottomrule
	\end{tabular}
 \label{tab:real-exp}
 \vspace{-5mm}
\end{table}

%% 图五（√）
\begin{figure}
	\centering
	\includegraphics[width=0.95\linewidth]{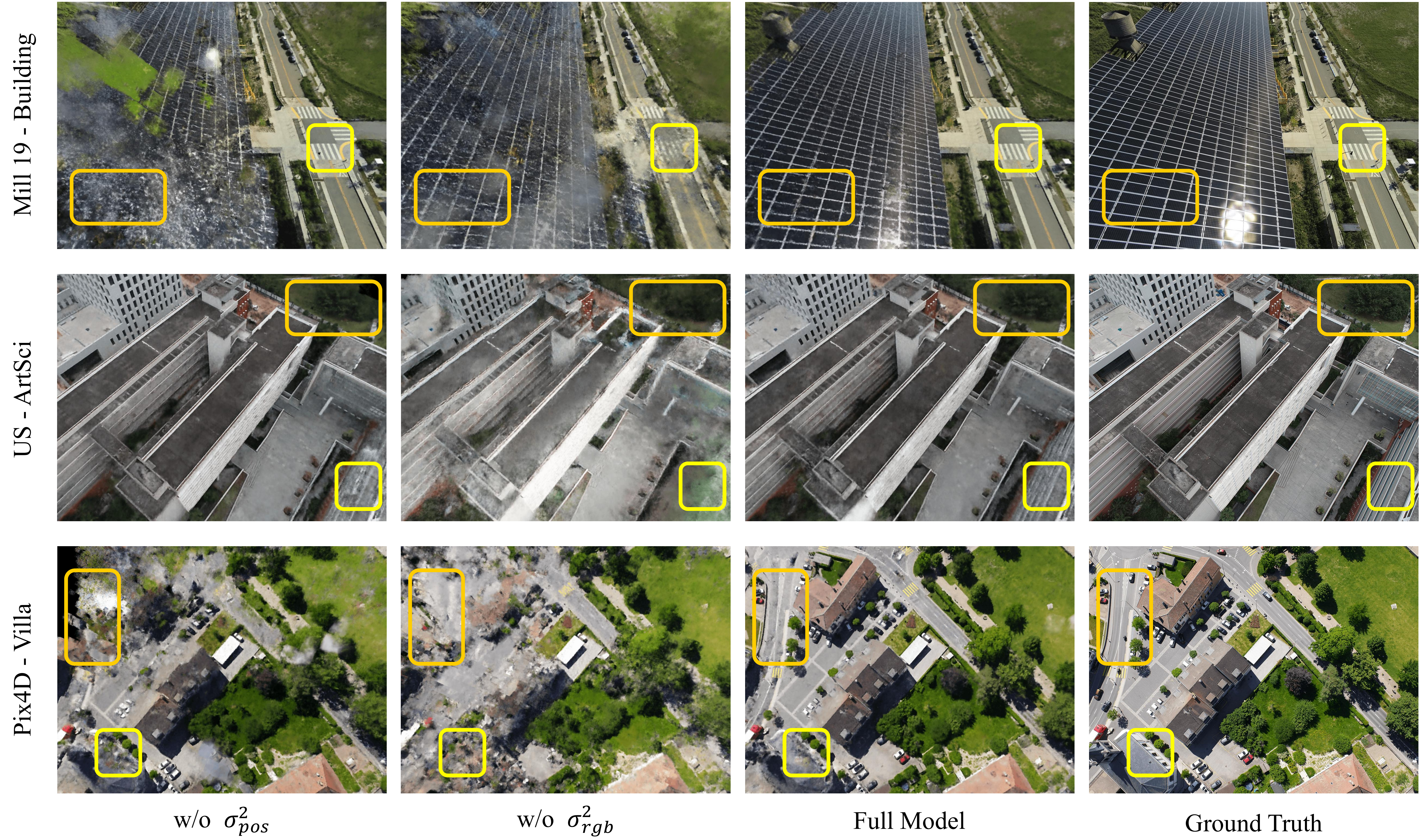}
	\caption{Rendering effect visualization of different components (partial experimental results).}
	\label{fig:ablation-exp}
    \vspace{-1mm}
\end{figure}

%% 表格二（√）
\begin{table}[t]
    \caption{Quantitative results of different components in ablation study(best shown in \textcolor{green}{green} and second in \textcolor{blue}{blue}).}
    \resizebox{\linewidth}{!}{
    \begin{tabular}{cccccccc}
        \toprule
        \diagbox{Component}{Scene} & & Building & Rubble & Polytech & ArtSci & Villa & University \\
        \midrule
        \multirow{3}{*}{w/o $\sigma _{pos}^2$} 
        & PSNR$\uparrow$ & 17.031 & 19.803 & \textcolor{blue}{19.856} & \textcolor{blue}{18.386} & \textcolor{blue}{15.937} & \textcolor{blue}{16.782}  \\
        & SSIM$\uparrow$ & \textcolor{blue}{0.404} & \textcolor{blue}{0.527} & \textcolor{blue}{0.541} & \textcolor{blue}{0.497} & 0.356 & 0.329 \\
        & LPIPS$\downarrow$ & \textcolor{blue}{0.407} & 0.645 & 0.731 & 0.693 & 0.645 & 0.669 \\
        \midrule
        \multirow{3}{*}{w/o $\sigma _{rgb}^2$} 
        & PSNR$\uparrow$ & \textcolor{blue}{17.841} & \textcolor{blue}{20.045} & 18.606 & 17.687 & 15.917 & 16.338 \\
        & SSIM$\uparrow$ & 0.390 & 0.491 & 0.534 & 0.451 & \textcolor{blue}{0.368} & 0.305 \\
        & LPIPS$\downarrow$ & 0.421 & \textcolor{blue}{0.341} & 0.534 & \textcolor{blue}{0.352} & \textcolor{blue}{0.475} & \textcolor{blue}{0.443} \\
        \midrule
        \multirow{3}{*}{Full Model} 
        & PSNR$\uparrow$ & \textcolor{green}{19.841} & \textcolor{green}{21.108} & \textcolor{green}{21.810} & \textcolor{green}{18.836} & \textcolor{green}{18.250} & \textcolor{green}{18.722} \\
        & SSIM$\uparrow$ & \textcolor{green}{0.469} & \textcolor{green}{0.573} & \textcolor{green}{0.592} & \textcolor{green}{0.500} & \textcolor{green}{0.457} & \textcolor{green}{0.430} \\
        & LPIPS$\downarrow$ & \textcolor{green}{0.351} & \textcolor{green}{0.303} & \textcolor{green}{0.235} & \textcolor{green}{0.319} & \textcolor{green}{0.386} & \textcolor{green}{0.327} \\
        \bottomrule
    \end{tabular}
    }
    \label{tab:ablation-exp}
    \vspace{-4mm}
\end{table}

%% Experiments第四段（√）
This comparison experiment demonstrates that our uncertainty estimation strongly correlates with novel view synthesis quality for NeRFs. As shown in Fig.~\ref{fig:real-experiment} and Table.~\ref{tab:real-exp}, IOVS4NeRF achieves significantly more information in uncertainty prediction with respect to synthesis error. The superior performance of our approach demonstrates the proposed hybrid uncertainty leads to more consistent uncertainty estimates.

%\subsection{Ablation Study}\label{ITH}
%% Experiments第五段（√）
%This ablation study is to further evaluate the influence of the individual components of our hybrid uncertainty. As shown in Fig.~\ref{fig:ablation-exp} and Table.~\ref{tab:ablation-exp}, IOVS4NeRF obviously outperforms ablated approaches consistently which indicates that both rendering uncertainty $\sigma _{rgb}^2$ and positional uncertainty $\sigma _{pos}^2$ contributes to the uncertainty estimation of the candidate view and improve the capability of informative view selection.

\subsection{Ablation Study}\label{ITH1}
%% Experiments第五段（√）

The ablation study aims to assess the impact of the individual components within our hybrid uncertainty model. The result, as depicted in Fig.~\ref{fig:ablation-exp} and Table.~\ref{tab:ablation-exp}, demonstrates that omitting the positional uncertainty component, $\sigma_{pos}^2$, leads to high-fidelity rendering in localized areas. However, this setting neglects broader scene coverage, as the next view selection is biased towards the unrendered regions. Conversely, disregarding the rendering uncertainty component, $\sigma_{rgb}^2$, results in a globally low-fidelity rendering. This occurs due to the next view selection overextending its reach, attempting to encompass the entire spatial extent. In contrast, our proposed hybrid uncertainty consistently produces superior outcomes by employing the combined measure of $\sigma_{rgb}^2$ and $\sigma_{pos}^2$. This indicates the synergistic benefits of incorporating both components into the uncertainty estimation process.
%As shown in Fig.~\ref{fig:ablation-exp} and Table.~\ref{tab:ablation-exp}, ignoring $\sigma _{pos}^2$ results in a localized high-fidelity rendering, but more regions are not rendered due to that the next view selection focused on the unrendered parts of scenes. In contrast, ignoring $\sigma _{rgb}^2$ achieves global low-fidelity rendering because the next view selection is overly ambitious in covering the full extent of space.

%% In contrast, our proposed hybrid uncertainty consistently produces superior outcomes by employing the combined measure of $\sigma_{rgb}^2$ and $\sigma_{pos}^2$. This indicates the synergistic benefits of incorporating both components into the uncertainty estimation process.
%However, using the sum of $\sigma _{rgb}^2$ and $\sigma _{pos}^2$ as the proposed hybrid uncertainty consistently yields the best results, indicating the complementarity of both components.

\section*{Conclusion}
In this paper, we propose an incremental optimal view selection framework, IOVS4NeRF, aiming at addressing the limitations of NeRF in resource-constrained scenarios. By expanding the training dataset with strategically selected samples based on a novel hybrid uncertainty, which takes into account both rendering uncertainty and positional uncertainty of different flight trajectories, IOVS4NeRF significantly enhances the quality of novel view synthesis while minimizing additional resource requirements. The efficacy has been validated through extensive experiments on large-scale realistic scenes, demonstrating its robustness even with limited training data. In future work, our proposed method could be extended to accommodate a broader range of scale-varying scenarios.

%\section*{Acknowledgement}
%This project is supported by Provincial Natural Science Foundation of Hunan(2024JJ10027).

\bibliography{main.bib}

\end{document}